\pdfoutput=1
 
\documentclass[11pt]{article}

\usepackage[]{emnlp2021}

\usepackage{times}
\usepackage{epsfig}
\usepackage{graphicx}
\usepackage{amsmath}
\usepackage{amssymb}
\usepackage{booktabs}
\usepackage{multirow}
\usepackage{multicol}
\usepackage{bm}
\usepackage{colortbl}
\definecolor{na}{gray}{0.9}

\usepackage{color,soul}
\usepackage[caption=false]{subfig}
\usepackage{microtype}      

\usepackage{algorithm}
\usepackage{algpseudocode}%

\usepackage{times}
\usepackage{latexsym}

\usepackage[T1]{fontenc}
\usepackage{subfig}

\usepackage[utf8]{inputenc}

\usepackage{microtype}
\usepackage{bm}
\newcommand{\setupfull}{Continual LearnIng of Few-shot Learners}
\newcommand{\setup}{CLIF}
\newcommand{\method}{BiHNet}
\newcommand{\eg}{\textit{e.g.}}

\title{
Learn Continually, Generalize Rapidly: \\
Lifelong Knowledge Accumulation 
for Few-shot Learning}

\author{Xisen Jin\raisebox{3pt}{$\S$} \quad Bill Yuchen Lin\raisebox{3pt}{$\S$} \quad Mohammad Rostami\raisebox{3pt}{$\dagger$}  \quad Xiang Ren\raisebox{3pt}{$\S$} \\
\raisebox{1pt}{$\S$}University of Southern California,~~
\raisebox{1pt}{$\dagger$}Information Sciences Institute\\ 
\texttt{\{xisenjin, yuchen.lin, xiangren\}@usc.edu} ~~
\texttt{\{mrostami\}@isi.edu} \\
}

\begin{document}
\maketitle
\begin{abstract}

The ability to continuously expand knowledge over time and utilize it to rapidly generalize to new tasks is a key feature of human linguistic intelligence. Existing models that pursue rapid generalization to new tasks (\eg, few-shot learning methods), however, are mostly trained in a single shot on fixed datasets, unable to dynamically expand their knowledge; while continual learning algorithms are not specifically designed for rapid generalization. 
We present a new learning setup, Continual Learning of Few-Shot Learners (CLIF), to address the challenges of both learning settings in a unified setup. 
CLIF assumes a model learns from a sequence of diverse NLP tasks arriving sequentially, accumulating knowledge for improved generalization to new tasks, while also retaining performance on the tasks learned earlier. 
We examine how the generalization ability is affected in the continual learning setup, evaluate a number of continual learning algorithms, and propose a novel regularized adapter generation approach. 
We find that catastrophic forgetting affects generalization ability to a lesser degree than performance on seen tasks; while continual learning algorithms can still bring considerable benefit to the generalization ability\footnote{Code and data are publicly available at \url{https://github.com/INK-USC/CLIF}}.

\end{abstract}


\section{Introduction}


The ability to recall acquired knowledge for  learning new tasks quickly and efficiently over time has been seen as a crucial metric of general linguistic intelligence~\cite{Yogatama2019LearningAE}.
Progress on this research problem has led to remarkable improvements in recent works on few-shot learning ~\cite{Brown2020LanguageMA, Gao2020MakingPL}. 
However, these methods have primarily focused on learning from a \textit{static} set of tasks (datasets) in an \textit{offline} manner, without \textit{dynamically} expanding the acquired knowledge over time. This training scheme is in contrast with the way humans process natural language~\cite{chomsky2002syntactic,montague1970universal}: humans are able to process novel meanings by retaining past knowledge, combining/decomposing chunks of language into prior learned language components, and avoid learning from scratch. 

Motivated by this observation, we study   whether NLP models could accumulate generalizable knowledge continuously over a sequence of tasks and learn to generalize to new tasks rapidly (\textit{i.e.}, with few examples).
This problem has not been investigated in the existing works --- a related line of efforts that look to learn from sequentially arriving tasks, known as continual learning (CL) or lifelong learning~\cite{robins1995catastrophic, Sun2020LAMOLLM, dAutume2019EpisodicMI}, mainly focus on retaining the performance on seen tasks when the model is continuously updated on new tasks (\textit{i.e.}, to overcome the catastrophic forgetting issue). 


\begin{figure}
    \centering
    \includegraphics[width=\linewidth]{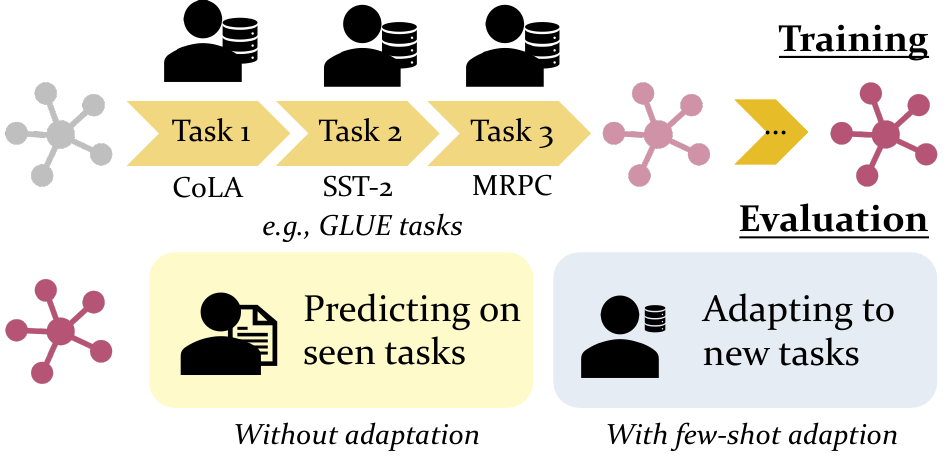}
    \caption{\textbf{Overview of the Training and Evaluation setup in \setup.} The model learns over a number of training tasks sequentially and is evaluated over all the seen tasks. We also evaluate its ability to adapt to new tasks with only a small number of labeled examples.}
    \label{fig:intro}
\end{figure}



To study this ability, we propose the~\setupfull~(\setup) setup (illustrated in Figure~\ref{fig:intro}) to simulate the challenge: In~\setup, the model learns over a sequence of NLP tasks (arriving one by one; without revisiting), and then evaluated in terms of (i) generalization to new (few-shot learning) tasks; and (ii) preserving its performance on solving seen tasks. We train and evaluate over a diverse set of NLP tasks, spanning over entity typing, sentiment analysis, natural language inference, and other classification tasks. 

With the \setup~setup, we conduct a series of experiments on existing models, in order to understand the relationship between continuous knowledge accumulation and few-shot generalization.
Our first analysis is to understand how the generalization ability evolves during continual training, and whether catastrophic forgetting affects the acquisition of generalization ability. We find a negative effect of catastrophic forgetting on the generalization ability, and a stronger negative effect on the performance over the seen tasks. 

In a follow-up analysis, we find most existing CL methods hardly benefit models' generalization ability, even they are shown to alleviate catastrophic forgetting. This implies some non-trivial challenges for accumulating knowledge that can help model generalization.
Inspired by recent research on Hypernetworks for few-shot learning~\cite{Requeima2019FastAF} and   continual learning approach using Hypernetworks~\cite{Oswald2020ContinualLW}, we propose Bi-level Hypernetworks for Adapters with Regularization to address challenges of the~\setup. We evaluate these approaches extensively by varying the number of training examples and the orders of tasks at training.

To summarize, the main contribution of this work is threefold (1) we propose CLIF setup, its data streams and protocols to comprehensively evaluate lifelong knowledge accumulation in NLP, and (2) we compare existing algorithms to demonstrate weaknesses of these algorithms (3) and propose Bi-level Hypernetworks for Adapters with Regularization as a solution to inspire future works.




\section{Problem Formulation}


\subsection{The CLIF Problem}
\label{ssec:problem_definition}
We assume there is an NLP model $f$ trained \textit{continually} on different tasks over time (i.e.,  continual learning), and then \textit{rapidly} generalizes to many unseen tasks with few-shot examples (i.e., few-shot adaptation).
In the \textit{continual learning} stage,
the model encounters an \textit{ordered} list of $N_u$ \textit{u}pstream tasks: $[\mathcal{T}_u^1, \dots, \mathcal{T}_u^{N_u}]$, where each task has its own training and test sets.
To test the few-shot learning ability of the sequentially trained model $f$,
we then adapt it on a set of $N_v$ \textit{few-shot} tasks individually $\{\mathcal{T}_v^i\}_{i=1}^{N_v}$, where only a few training examples are available for each unseen task.
We name this learning setting as \textsc{CLIF}, which stands for continual learning for few-shot adaptation. In addition to the traditional   objective in CL to preserve performance on seen tasks, in \textsc{CLIF} it is also crucial to retain generalizable knowledge to achieve better few-shot learning performance at the end of training.



%

\paragraph{Evaluation Protocol}
\begin{figure}
    \centering
    \includegraphics[width=\linewidth]{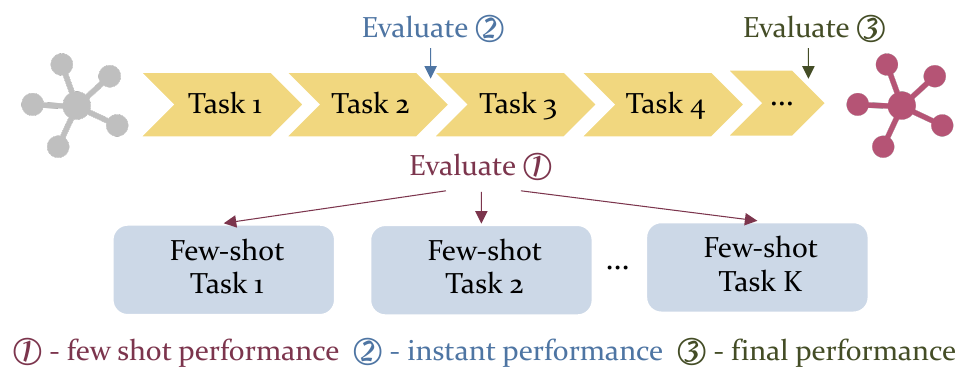}
    \caption{\textbf{Evaluations setups in~\setup}. (1) and (2) measure generalization ability to new tasks; while (3) indicate forgetting on seen tasks.}
    \label{fig:eval}
\end{figure}
As illustrated in Figure~\ref{fig:eval},
there are three major aspects for evaluating a method to the CLIF setting: few-shot performance, final performance, and instant performance.

\smallskip
\noindent
1) \textbf{\textit{Few-shot Performance}}. First, we evaluate the continually trained model $f$ on a set of unseen tasks, by fine-tuning it for each task $\mathcal{T}_v^i$ individually with a few annotated examples when the training over upstream tasks $\mathcal{T}_u^1..\mathcal{T}_u^{N_u}$ ends. Thus, we can assess the \textit{ few-shot generalization} ability. We note the few-shot accuracy for a task $\mathcal{T}_v^i$ as $s_{\text{FS}}^i = F(\mathcal{Y}_v^i,\hat{\mathcal{Y}}_v^{i}) $, where $\hat{\mathcal{Y}}_v^{i}$ is the predictions over the test examples of task $\mathcal{T}_v^i$, $\mathcal{Y}_v^i$ is the set of ground truth labels, and $F$ is the metric function (\eg, accuracy). We report $s_{\text{FS}}$ averaged over all few-shot tasks, \textit{i.e.,} $s_{\text{FS}} = \frac{1}{N_v} \sum_{i=1}^{N_v} s_{\text{FS}}^i$. We also compute a relative improvement $\Delta_{\text{FS}}=\frac{s_{\text{FS}} - s^\prime_{\text{FS}}}{s^\prime_{\text{FS}}}$ over the performance $s^\prime_{FS}$ of the models separately trained on each few-shot task.

\smallskip
\noindent
2) \textbf{\textit{Instant Performance}}. We evaluate the performance of an upstream task $\mathcal{T}_u^i$ right after the model $f$ finishes the learning on it. We note the set of model prediction on the test set of task $\mathcal{T}_u^i$ right after the model $f$ learns the task $j$ as  $\hat{\mathcal{Y}}_u^{i,j}$. The instant performance over task $\mathcal{T}_u^i$ is defined as $s_{\text{inst.}}^i = F(\mathcal{Y}_u^i,\hat{\mathcal{Y}}_u^{i,i})$.
For example, we evaluate the performance of $f$ on $\mathcal{T}_u^2$ after the model $f$ is trained on the data of $\mathcal{T}_u^1$ and $\mathcal{T}_u^2$, before further train it on $\mathcal{T}_u^3$.
The performance of $f$ on $\mathcal{T}_u^2$ now can thus tell us how well the model \textbf{\textit{transfers}} its knowledge from learning $\mathcal{T}_u^1$ to learn $\mathcal{T}_u^2$ --- using the performance when $f$ is trained only on $\mathcal{T}_u^2$ as a reference.  We compute average instant performance of all upstream tasks, $s_{\text{inst.}} = \frac{1}{N_u} \sum_{i=1}^{N_u} s_{\text{inst.}}^i$ We additionally compute a relative improvement $\Delta_{\text{Inst.}} = \frac{s_\text{inst.} - s^\prime_\text{inst.}}{s^\prime_\text{inst.}}$ over the performance $s^\prime_{\text{inst.}}$ of models separately trained on each upstream task to indicate benefit of upstream learning.

\smallskip
\noindent
3) \textbf{\textit{Final Performance}.} We also evaluate the performance of $f$ at the end of the continual learning over upstream tasks to know how much the model $f$ \textbf{\textit{forgets}} the knowledge about the task after it learns to solve more tasks. The final accuracy $s_{\text{final}}^i$  of a task $\mathcal{T}_u^i$ is defined as $F(\mathcal{Y}_u^i,\hat{\mathcal{Y}}_u^{i,N_u})$. Similarly, we report the averaged final accuracy over all tasks, noted as $s_{\text{final}} = \frac{1}{N_u} \sum_{i=1}^{N_u} s_{\text{final.}}^i$. For a single model, the forgetting can be quantified as $s_{\text{inst}} - s_{\text{final}}$.

\paragraph{Challenges}
The CLIF setting is particularly challenging for existing few-shot learning methods. 
Most few-shot learning methods assume that   the upstream training datasets for all tasks are always available and there is no temporal order for learning. Hence, the    upstream tasks can be learned jointly in a multi-task learning setting.
However, the CLIF problem follows a \textit{continual learning} setup, where the tasks are visited sequentially without revisiting.
Thus, methods relying on random sampling from a task distribution are not applicable.




\subsection{Tasks and Data Streams}
\label{ssec:datasets}
\begin{table}[t]
\centering
\scalebox{0.7}{
\begin{tabular}{ rp{5.2cm}c }
\toprule
 Learning Stage & \multicolumn{1}{c}{Tasks} & \# Tasks \\ \midrule
\multicolumn{3}{c}{\textbf{\texttt{CLIF-26}} } \\
Continual ($\mathcal{T}_u$) &    GLUE~\cite{wang2019glue}       &  $N_u=$9  \\
Few-shot ($\mathcal{T}_v$) & DivFSL~\cite{Bansal2020LearningTF}  & $N_v=$17    \\
\midrule
\multicolumn{3}{c}{\textbf{\texttt{CLIF-55}} }   \\
Continual  ($\mathcal{T}_u$)   &    SuperGLUE-RTE, TweetEval-Sentiment, Scicite, GLUE-MRPC, Scitail, KILT-Fever, ...     &  $N_u=$45 \\
Few-shot ($\mathcal{T}_v$) &    SuperGLUE-CB, Dbpedia-14, Wiki-QA, emo, Yelp-Polarity, ethos-religion, tab-fact, financial-phrasebank,  ANLI, ethos-race  &  $N_v=$10    \\
\bottomrule
\end{tabular}
}
\caption{\textbf{Overview of datasets employed for upstream continual training and few-shot learning}. We include the full list of tasks in Appendix~\ref{sec:apdx_imp}.}
\label{tab:datasets}
\end{table}

To push the~\setup~challenge to a more practical setup, we consider a diverse set of NLP tasks to perform CL and few shot learning. We consider two dataset combinations, referred to as \texttt{CLIF-26} and \texttt{CLIF-55} tasks, summarized in Table~\ref{tab:datasets}. In the first combination, following~\citet{Bansal2020LearningTF}, we use the GLUE~\cite{wang2019glue} benchmark as our upstream tasks for CL stage for experiments which consists of $N_u=9$ tasks. We then evaluate the few-shot learning ability over $N_v=17$ DivFSL~\cite{Bansal2020LearningTF} tasks, spanning over diverse NLP tasks including sentiment analysis, entity typing and natural language inference.
In \texttt{CLIF-55}, we train and test the model over $N_u=45$ and $N_v=10$ tasks selected from Huggingface datasets library\footnote{\url{https://huggingface.co/datasets}}. The selected datasets span over a broad family of NLP tasks, including natural language inference, emotion classification, topic classification, fact checking, hate speech detection, paraphrasing, and others.


To adopt it for our learning setting, we specify an order on the tasks presented to the model for \texttt{CLIF-26} and \texttt{CLIF-55} (details in Appendix~\ref{sec:apdx_imp}).
We also consider alternative task orders in our experiments. The model sequentially visits each task during training.  We limit the number of training examples in each GLUE task in \texttt{CLIF-26} to 10,000 to avoid overly imbalanced datasets. For \texttt{CLIF-55}, we use 90 examples per class for continual learning. We use $k=16$ examples per class in few-shot learning tasks for both~\texttt{CLIF-26} and~\texttt{CLIF-55} if not specified, and include more setups of $k$ in the experiments.
As the test labels for GLUE are not publicly available, we report performance on validation sets. We convert regression tasks (\textit{e.g.} STS-B) to binary classification tasks by setting the threshold in the middle of the maximum and minimum regression scores.

All examples are converted into sequence-to-sequence question-answering formats following~\cite{mccann2018natural} to allow a single model to solve all tasks. We consider \textit{exact match} between the generated answer span and the ground-truth span as a correct prediction.  
For both the upstream tasks and few-shot tasks in $\texttt{CLIF-26}$ and $\texttt{CLIF-55}$, we use the prediction \textit{accuracy} as the metric function.

\section{Method}

This section presents baseline methods to set up the lower bounds for the CLIF problem, and approaches to improve the performance. We view an approach by its \textit{base model} and the \textit{learning algorithm}. We first introduce the base models in our study (Sec.~\ref{ssec:basearch});
Then, we introduce a few existing methods for continual learning and continual meta-learning (Sec.~\ref{ssec:baseline}). 
Finally, we present a novel regularized bi-level adapter generation framework to better address the CLIF problem (Sec.~\ref{ssec:method}).



\subsection{Base NLP Models}
\label{ssec:basearch}
\paragraph{BART and BART-Adapter.} As we formulate the NLP tasks in the CLIF problem in a unified text-to-text format, we use pre-trained language models (LMs) as the architecture of the model $f$ and fine-tune the entire model during training.
We mainly use the BART-base~\cite{lewis-etal-2020-bart} model for 
our experiments.
We also include Adapter training~\cite{Houlsby2019ParameterEfficientTL} as an alternative to fine-tuning the entire BART model.
Here, \textit{adapters}~\cite{Houlsby2019ParameterEfficientTL} are two-layer Multi-Layer Perceptrons (MLPs) plugged after each layer of BART. Given the output $h_\ell$ at the $\ell$-th layer of the transformer, the adapted output is computed as $h_\ell^\prime = h_\ell + f^a_\ell(h_\ell)$, where $f^a_\ell$ is the adapter layer at layer $\ell$. 
Only adapters are learned during training, while the BART model is   frozen. We note two approaches \texttt{BART} and \texttt{BART-Adapter} respectively.

\paragraph{Hyper-Networks for Adapter Generation.} In addition to BART and BART Adapter, we also use consider a HyperNetwork (\texttt{HNet}) architecture. The hypernetwork, noted as $g$, takes a task representation $z$ as input and generates model parameter of another prediction model, noted as $f$ to solve the task. In few-shot learning, $z$ is usually computed as the average representation of training examples of the task, $ \bm{z} = \frac{1}{|\mathcal{D}_{tr}^i|} \sum_{(\bm{x}_j, \bm{y}_j) \in \mathcal{D}_{tr}^i} f_e(\bm{x}_j, \bm{y}_j)$, where $\mathcal{D}_{tr}^i$ is the training set of the task $\mathcal{T}^i$ and $f_e$ in an encoder model. In our case, we use a BART model as $f_e$ and feed it the concatenation of $\bm{x}$ and label $\bm{y}$ in text format to obtain the task representation $z$. As the model allows flexible control of model parameters with training examples, it is broadly applied for few-shot learning~\cite{Requeima2019FastAF, Gidaris2018DynamicFV}; besides, $\bm{z}$ can also be randomly initialized and end-to-end learned~\cite{ha2016hypernetworks}. As the parameter space of large-scale PTLMs like BART is huge, following~\cite{Ye2021ZeroshotLB}, we generate model parameters only for adapters.

In summary, we consider \texttt{BART} fine-tuning, \texttt{BART-Adapter} learning and \texttt{HNet} for adapter generalization as three base NLP models. In   section~\ref{ssec:baseline}, we introduce algorithms to learn these models in the \setup~setting.



\subsection{Baseline Learning Algorithms}
\label{ssec:baseline}



\paragraph{Single Task Learning} 
To understand the reference performance of a base model on an \textit{\textbf{upstream}} task without any knowledge transfer, we apply the single task learning (STL) method,
which trains and tests a model $f$ on the dataset of each task in isolation.
In this case, we ignore the sequential nature of the CLIF problem so we can use this STL performance to assess the effectiveness of different continual methods (introduced below). 
Ideally, a valid CL algorithm should have a better few-shot accuracy than STL results, meaning that it accumulates knowledge and effectively transfer it for learning.  
Similarly, to know the reference performance of the \textit{\textbf{few-shot}} tasks, we learn a model $f$ for each few-shot task on the given examples, \textit{without} any upstream training, so that we can use such performance to assess how well a CLIF method  improves the generalization ability. 



\paragraph{Continual Learning Algorithms} 
As a straightforward baseline method,
we use \texttt{Vanilla} to denote simply training the model $f$ sequentially on the upstream tasks.
Specifically, it trains the model $f$ on $\mathcal{T}_u^{i}$ until its performance converges and then continually train $f$ on the data of $\mathcal{T}_u^{i+1}$.
Note that the access of the data on previous tasks is not allowed in CL. We also consider CL algorithms such as \texttt{EWC}~\cite{kirkpatrick2017overcoming}, \texttt{MbPA++}~\cite{dAutume2019EpisodicMI} and \texttt{meta-MbPA}~\cite{Wang2020EfficientML} in our experiments. We use an online variant of EWC~\cite{schwarz2018progress}.
EWC regularizes the change of important model parameters during training.
The \texttt{MbPA++} method performs test-time adaptation over a few training examples stored in the memory. 
The \texttt{meta-MbPA} method includes a meta-learning objective to adapt fast. 

As a comparator that does not suffer from forgetting, we also report the results of \textbf{multi-task learning} over upstream tasks (\texttt{MTL}) for reference.

\paragraph{Hyper-Networks for CL.}~\citet{Oswald2020ContinualLW} proposed a hypernetwork-based continual learning algorithm, where the high-level idea of mitigating catastrophic forgetting is to penalize the hypernetwork for the change of generated model weights for previous tasks when it learns a new task. While the original work generates entire parameters of a model, we adapt it to PTLMs by generating the weights of adapters only. We note the approach as \texttt{HNet-Reg}. 

Specifically, when the model has just finished learning the task $\mathcal{T}_u^{i-1}$  and right before learning the task $\mathcal{T}_u^i$ in the continual learning stage,
we compute the adapter weights generated by our \textit{current} hypernetwork for all prior tasks $\mathcal{T}_u^1..\mathcal{T}_u^{i-1}$, noted as $\{\hat{\bm{\theta}}_{1}^{i-1}, \hat{\bm{\theta}}_{2}^{i-1}, \dots, \hat{\bm{\theta}}_{i-1}^{i-1}\}$ --- where the generation is controlled by applying the hypernetwork $h$ on the stored task representations of previous tasks $1..i-1$, noted as $\mathcal{M}=\{\bm{z}^{1}_h, \dots, \bm{z}^{i-1}_h\}$. Here, the task representation $\bm{z_i}$ for task $\mathcal{T}_u^i$ is randomly initialized before learning the task and optimized jointly while learning the task.
Then, in each step of learning $\mathcal{T}_u^i$, we randomly sample a prior task $\mathcal{T}_u^j$ ($j<i$) to regularize the hypernetwork learning.
It penalizes the $\ell_2$ distance between the adapter weights generated at the current step $\bm{\theta}_{j}$ and the pre-computed one, \textit{i.e.}, $\lvert \lvert \bm{\theta}_{j} - \hat{\bm{\theta}}^{i-1}_{j} \rvert \rvert_2^2$.
Therefore, we avoid the hypernetwork $g$ changes its output for a prior task too much during the continual learning stage, so that the knowledge accumulation is better guaranteed for the learned model.

\paragraph{Limitations.}  EWC and HNET-Reg are not well-designed for the CLIF problem, which additionally tries to improve the few-shot generalization on \textit{unseen} tasks after continual learning. While the test-time adaptation in MbPA and meta-MbPA may benefit few-shot learning, such ability is not studied in these works. Besides, as these two algorithms store \textit{real examples} of previous training tasks, it is not applicable in privacy sensitive applications where data from earlier task is no longer accessible, which is a typical scenario in continual learning.



\begin{figure}
    \centering
    \includegraphics[width=\linewidth]{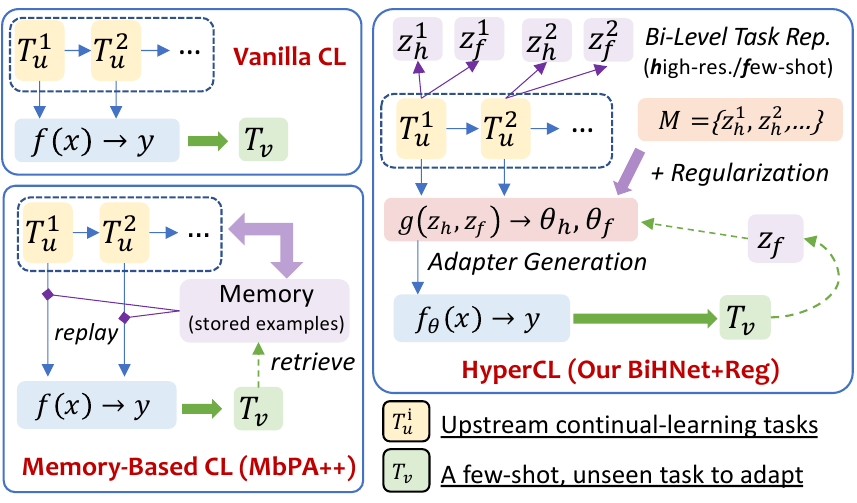}
    \caption{
    \textbf{A comparison between different typical continual methods to the CLIF problem}. The Vanilla CL method simply trains the model on a sequence of tasks $\mathcal{T}_u$. Memory-based methods such as MbPA++~\cite{dAutume2019EpisodicMI} store a small set of examples of prior tasks and then replay them during learning. Our BiHNet+Reg method uses a hypernetwork to generate the weights of model adapters according to bi-level (high-resource and few-shot) task representations.
    }
    \label{fig:architecture}
\end{figure}

\subsection{Our Extension: Bi-level Hypernetworks for Adapters with Regularization}
\label{ssec:method}
Inspired by hypernetwork approaches for few-shot learning and continual learning,  we extend the hypernetwork-based CL methods for CLIF.
We present a novel method, Bi-level Hypernetwork for Adapters with Regularization (\texttt{\method+Reg}), 
which learns to use the bi-level task representations to generate adapter weights for learning
a fast adaptive model over a sequence of tasks, while mitigating the forgetting effect via regularization. 

As shown in Figure~\ref{fig:architecture}, the proposed method consists of three components: 
(1) a context predictor to generate bi-level task representations (i.e., high-resource and few-shot representations) from training examples, 
(2) a hypernetwork to generate weights of adapters given the task representations,
and (3) a regularization term to discourage weight changes of seen tasks to avoid forgetting following~\cite{Oswald2020ContinualLW}.
We discuss each individual component below.


\paragraph{Context Predictor.}

We propose to generate two task representations for each task $t$
to model it in the \textit{high-resource} and \textit{few-shot} cases respectively, denoted as $\bm{z}^t_h$ and $\bm{z}^t_f$,  with a frozen BART model.
The high-resource representations are used to encourage the knowledge transfer during continual learning; the few-shot task representations help us mimic the few-shot tasks in the few-shot learning stage for better generalization, similar to meta-learning. Specifically, we use an LM (e.g., BART) as the context representation model $R$ for encoding an example $(\bm{x}, \bm{y})$: we feed $\bm{x}$ and $\bm{y}$ to the encoder and the decoder of the model $R$, and use the latent representation from this last-layer activation.
The \textbf{high-resource} task representation is then computed as the average of \textbf{all} examples' representations in task $t$, noted as $ \bm{z}^t_h = \frac{1}{|\mathcal{D}_t|} \sum_{(\bm{x}_i, \bm{y}_i) \in \mathcal{D}_t} R(\bm{x}_i, \bm{y}_i)$; while the few-shot task representation $\bm{z}^t_f$ uses the average of a limited number (say, $K$) of \textbf{sampled} examples $\bm{z}^t_f =\frac{1}{K} \sum_{(\bm{x}_i, \bm{y}_i) \in \Gamma(\mathcal{D}_t, K)}  R(\bm{x}_i, \bm{y}_i)$, where $\Gamma(\mathcal{D}_t, K)$ means sampled $K$ examples in  $\mathcal{D}_t$. 


Note that the high-resource representations of upstream tasks are stored in a memory module over time during the continual learning, $\mathcal{M}=\{\bm{z}^{t}_h | t \in \{\mathcal{T}^i_u\}_{i=1}^{N_u} \}$ . 
In the few-shot learning stage, we set $K$ as the number of given examples, so the $\bm{z}_h = \bm{z}_f$ for any tasks.



\paragraph{Adapter-Wise Hypernetworks.} 

Following the practice introduced in Sec.~\ref{ssec:basearch}, we use a hypernetwork $g$ to generate weights of adapters between the layers of the frozen BART model $f$. During training, we use high-resource and sampled task representations $z^t_h$ and $z^t_f$ to generate adapter weights separately, noted as $\bm{\theta}_t^h$ and $\bm{\theta}_t^f$. We optimize the prediction loss with both adapters.

\begin{table*}[t]
\centering
\scalebox{0.70}{
\begin{tabular}{@{}c||ccc|cc||ccc|cc@{}}
CLIF Dataset & \multicolumn{5}{c||}{\textbf{\texttt{CLIF 26}} (GLUE $\rightarrow$ DivFSL)} & \multicolumn{5}{c}{\textbf{\texttt{CLIF 55}} (Classification)} \\ \toprule
Methods $\downarrow$  Metrics $\rightarrow$ & Final Acc. & Inst. Acc. & F-S Acc. & $\Delta_{\text{Inst.}}$  & $\Delta_{\text{FS.}}$  & Final Acc.  & Inst. Acc.  & F-S Acc. & $\Delta_{\text{Inst.}}$  & $\Delta_{\text{FS.}}$  \\ \midrule
\rowcolor{na} \multicolumn{11}{c}{Single Task-Learning} \\\midrule
BART-Single & - & 79.39$_{\pm{0.7}}$ & 60.99$_{\pm{0.5}}$ & - & - & - & 69.32$_{\pm{0.3}}$  & 68.49$_{\pm{0.7}}$ & - & - \\
BART-Adapter-Single & - & 74.98$_{\pm{0.7}}$ & 59.00$_{\pm{1.9}}$ & - & - & - & 65.15$_{\pm{0.5}}$ & 65.70$_{\pm{0.8}}$ & - & -  \\
BiHNet-Single & - & 76.67$_{\pm{0.4}}$ & 52.66$_{\pm{0.9}}$ & - & - & - & 66.44$_{\pm{0.2}}$ & 64.57$_{\pm{1.1}}$ & -  & -  \\ 
\midrule
\rowcolor{na} \multicolumn{11}{c}{Continual learning} \\\midrule
BART-Vanilla & 19.73$_{\pm{0.2}}$ & 79.92$_{\pm{0.2}}$ & 58.96$_{\pm{3.2}}$ & 0.7\% & -3.3\% & 49.46$_{\pm{1.7}}$ & 71.26$_{\pm{0.6}}$ & 66.08$_{\pm{0.6}}$ & 2.8\% & -3.5\% \\
BART-MbPA++ & 59.52$_{\pm{1.0}}$ & 77.48$_{\pm{0.5}}$ & 56.26$_{\pm{1.4}}$ & -2.4\% & -7.8\% & 51.75$_{\pm{1.5}}$  & 67.18$_{\pm{1.0}}$ & 61.03$_{\pm 3.5}$ & -3.1\% & -10.9\% \\
BART-meta-MbPA & 55.69$_{\pm{0.9}}$ & 78.63$_{\pm{0.5}}$ & 57.88$_{\pm{1.0}}$ & -0.9\% & -5.1\% & 51.55$_{\pm 2.3}$ & 67.92$_{\pm 1.2}$ & 61.30$_{\pm 2.0}$  & -2.0\% & -10.5\% \\
BiHNet-Vanilla & 53.15$_{\pm{2.1}}$ & 79.90$_{\pm{0.2}}$ & 58.76$_{\pm{1.6}}$ & 4.2\% & -0.4\% & 44.03$_{\pm{1.7}}$ & 70.97$_{\pm{1.6}}$ & 66.23$_{\pm{0.6}}$ & 6.8\%  & 0.8\%  \\
BiHNet-EWC & 56.15$_{\pm{1.6}}$ & 78.73$_{\pm{0.3}}$ & 58.36$_{\pm{1.7}}$ & 2.7\% & -1.1\% & 7.15$_{\pm{2.1}}$ & 72.43$_{\pm{1.0}}$ & 58.08$_{\pm{0.8}}$ & 9.0\%  &  -11.6\%\\
BiHNet-Reg & 77.22$_{\pm{1.1}}$ & 80.24$_{\pm{0.4}}$ & 60.09$_{\pm{1.1}}$ & 4.7\% & 1.8\% & 56.16$_{\pm{1.6}}$ & 73.04$_{\pm{0.6}}$ & 68.46$_{\pm{0.2}}$ & 9.9\% & 4.2\%   \\\midrule
\rowcolor{na} \multicolumn{11}{c}{Multi-Task Learning} \\ \midrule
BART-MTL & 74.07$_{\pm{0.4}}$ & - & 55.02$_{\pm{2.5}}$ & - & -9.7\% & 63.78$_{\pm{0.0}}$ & - & 70.20$_{\pm{0.4}}$ & - & 2.5\% \\
BiHNet-MTL & 78.20$_{\pm{0.3}}$ & - & 59.22$_{\pm{0.8}}$ & - & 0.4\% & 64.93$_{\pm{0.0}}$ & - & 66.40$_{\pm{3.6}}$ & - & 1.1\% \\
Majority & 55.22 & - & 47.04 & - & - & 52.74 & - & 59.52 & -  & - \\ \bottomrule
\end{tabular}
} 

\caption{Final accuracy (Final Acc.) and instant accuracy (Instant Acc.) over upstream tasks and accuracy over few-shot learning tasks (Few-shot Acc.) on \texttt{CLIF-26} and \texttt{CLIF-55} tasks. We compute relative improvement of instant accuracy ($\Delta_{\text{Inst.}}$) and few-shot accuracy ($\Delta_{\text{FS}}$) over zero-knowledge baselines (the better one between BART-Adapter-Single and \method-Single for \method, and BART-Single for BART approaches).\footnotemark
}
\label{tab:main_tab}
\vspace{-0.3cm}
\end{table*}

\paragraph{Regularization.} 

Given that the HyperNetwork is the only trainable part in our model, we impose regularization on generated adapters to mitigate forgetting following HNet+Reg introduced in~\ref{ssec:baseline}. While our~\method~is trained to generate adapters from both high-resource and low-resource task representations, we find it sufficient to only store and regularize outputs from high-resource task representations.

\paragraph{Summary and Highlights}
To sum up, our proposed method first generates bi-level task representations for  training adapter-wise hypernetworks with a regularization term dedicated for avoiding forgetting over time.
Unlike replay-memory based CL approaches (\textit{e.g.}, MbPA~\cite{dAutume2019EpisodicMI}), our method does not store any real training examples. 
Instead, it uses task representations for storing the memory, and thus allows the method to be applied in privacy-sensitive scenarios.



\section{Results and Analysis}



We address our two major research questions in this section: (1) how models accumulate generalizable knowledge over time in a CL setup compared to offline setups given potential catastrophic forgetting, and (2) whether continual learning approaches reduce catastrophic forgetting of both seen-task performance and generalizable knowledge. We experiment with various combinations of model architectures in~\ref{ssec:basearch} and learning algorithms~\ref{ssec:baseline}. We note a method by its model architecture and CL algorithm applied, \eg, BART-Vanilla, \method-EWC. We include details of implementation in Appendix~\ref{sec:apdx_imp}.

\subsection{Examining Knowledge Accumulation}
\label{ssec:exp_no_cl}



In this section, we present analysis of model's ability to acquire generalizable knowledge in offline and CL setup. We note~\method~methods, which correspond to learning to generate adapters, should be compared with~\method-Single and BART-Adapter-Single, which are zero-knowledge baselines that learns to generate or learn adapters from random initialization; similarly, BART methods should be compared with BART-Single. We focus on identifying challenges in~\setup, and leave discussions of methodology in the next subsection.  \footnotetext{Note that, for single-task learning baselines, ``Inst. Acc." column is used to refer to the averaged accuracy of individual models trained for each upstream task.}

\paragraph{Q1: Is knowledge from upstream tasks helpful for a model's few-shot generalization in offline and continual learning setups?} To answer the question, we compare the performance of MTL with learning separate models per few-shot task without learning upstream tasks. Table~\ref{tab:main_tab} summarizes the results. On both \texttt{CLIF-26} and \texttt{CLIF-55} datasets, we see \method-MTL could outperform zero-knowledge baselines in few-shot Acc. by 0.4\% and 1.0\%, which implies upstream tasks are helpful for few-shot generalization in standard offline learning setups. For BART models, we notice BART-MTL improves over BART-Single on \texttt{CLIF-55} datasets by 2.5\%. However, we notice the opposite for \texttt{CLIF-26}. Given that the entire BART parameters are optimized in these models, we hypothesize that BART-MTL may have suffered from the forgetting of knowledge in the pre-trained BART model itself; while in adapter and~\method~models, the BART model is frozen. Therefore, in the rest of the section, we focus more on~\method~approaches.

\begin{figure}
    \centering
    \includegraphics[width=0.95\linewidth]{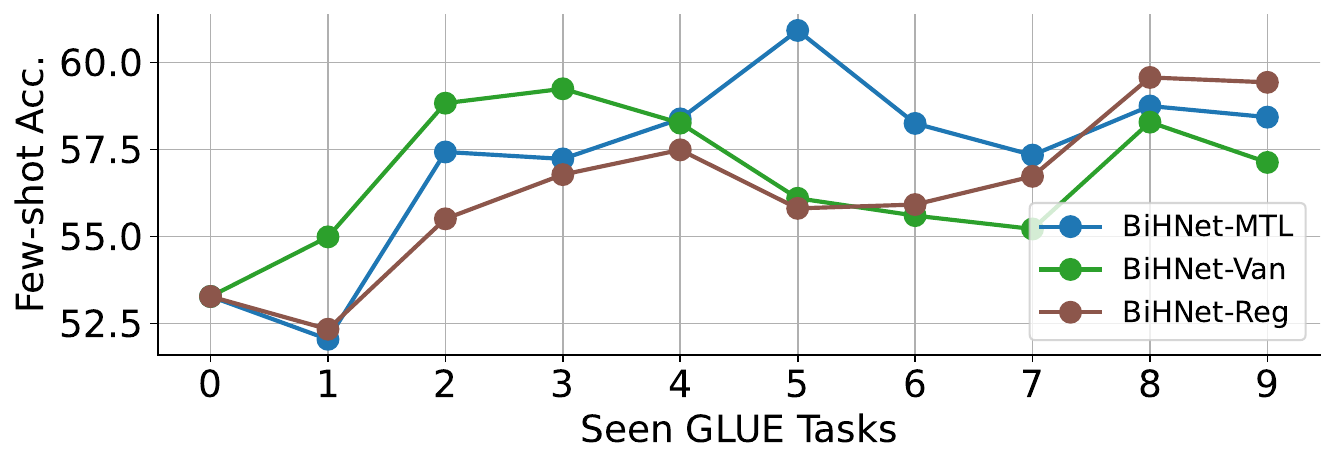}
    \caption{Few-shot learning performance on \texttt{CLIF-26} test tasks evaluated after each checkpoint of the model as the model sequentially visit upstream continual learning tasks.}
    \label{fig:leopard_fewshot_evolve}
\end{figure}


\paragraph{Q2: How does the model's generalization ability evolve over time?}
We focus on \method-Vanilla and BART-Vanilla approaches and answer three sub-questions.


\textit{Is the knowledge being monotonically accumulated over upstream tasks?}
In comparison to two zero-knowledge baselines, we notice \method-Vanilla generally improves both Instant Accuracy (4.2\% on \texttt{CLIF-26} and 6.8\% on \texttt{CLIF-55}) and Few-shot Accuracy (0.8\% on \texttt{CLIF-55}), except in few-shot Acc. on \texttt{CLIF-26} (-0.4\%). The results confirm positive knowledge accumulation to some extent.
In Figure~\ref{fig:leopard_fewshot_evolve}, we plot the few-shot accuracy on \texttt{CLIF-26} when the model sequentially visits each upstream training task. We note the few-shot accuracy of \method-Vanilla does not monotonically increase, which implies interference between these upstream learning tasks or forgetting of generalizable knowledge.

\begin{figure}
    \centering
    \includegraphics[width=0.8\linewidth]{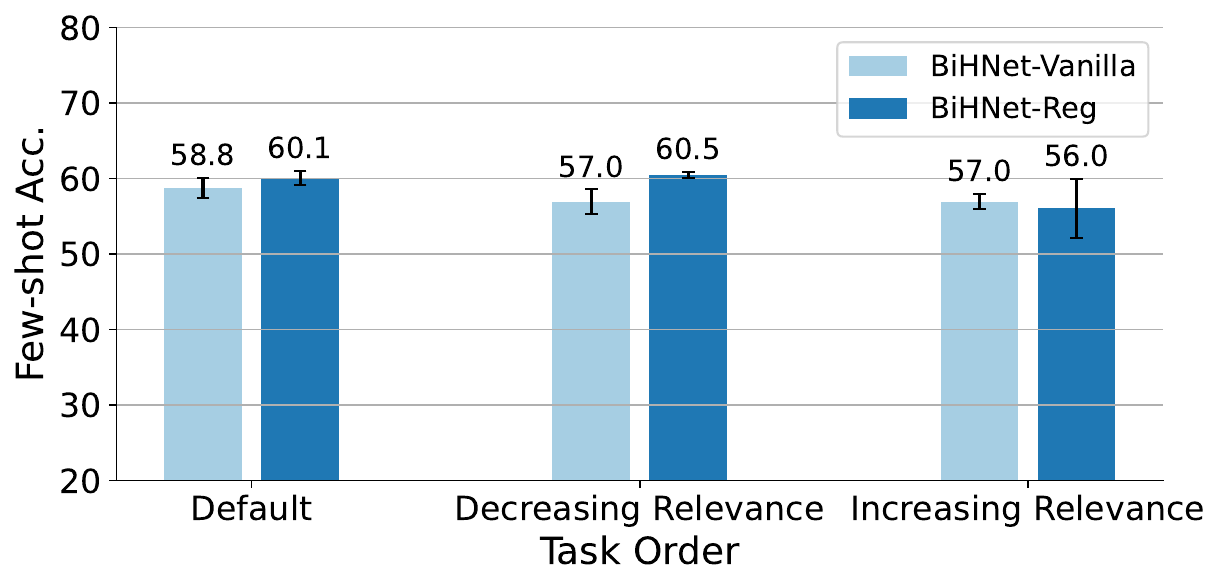}
    \caption{Few-shot learning performance of \method-Vanilla and \method-Reg on \texttt{CLIF-26} tasks when training tasks are presented in different orders.}
    \label{fig:few_shot_order_leopard}
\end{figure}

\textit{Does the order of the tasks matter?} 
Figure~\ref{fig:few_shot_order_leopard} present performance of methods under different orders of tasks on \texttt{CLIF-26}. We order the tasks by increasing and decreasing relevance to few-shot learning tasks, where the relevance is defined as few shot accuracy when the model transfers from a single upstream tasks. The results show in both orders \method-Vanilla is less competitive than BART-Adapter-Single. It implies that in continual learning the knowledge accumulation is less robust without CL algorithms.




\paragraph{Q3: Does model's catastrophic forgetting hinder its knowledge accumulation?} In Table~\ref{tab:main_tab}, we see clear differences between final accuracy of Vanilla and MTL approaches (by around 20 points), which verifies the catastrophic forgetting of seen-task performance when training examples are not i.i.d. However, we find the gap between MTL and Vanilla training is close for few-shot learning performance, where BART-Vanilla is even better than BART-MTL, which can be a positive outcome of adequate forgetting for alleviating over-fitting~\cite{Wang2020EfficientML}. It indicates the catastrophic forgetting influence generalization ability to a lesser degree compared to its effect on seen-task performance. 


\subsection{Effect of Continual Learning Algorithms} 
With the insights obtained for earlier questions, we now analyze whether baseline continual learning algorithms and the proposed approach help knowledge accumulation and improve models' (few-shot) generalization ability.

\paragraph{Q1: Do continual learning algorithms mitigate catastrophic forgetting?} From Table~\ref{tab:main_tab}, we notice MbPA++, meta-MbPA, EWC clearly improve final accuracy over BART-Vanilla or~\method-Vanilla on \texttt{CLIF-26}, which confirm positive effects on mitigating catastrophic forgetting. On~\texttt{CLIF-55}, which features much more training tasks and less examples per tasks, we find baseline CL algorithms fail to improve final accuracy. For memory-based approaches such as MbPA++ and meta-MbPA, it can because of significant overfitting to stored examples. In contrast, \method-Reg is effective in both datasets.

\paragraph{Q2: Does mitigating catastrophic forgetting better retain generalization ability?} On \texttt{CLIF-26}, by comparing the few-shot accuracy of \method-Vanilla and \method-Reg, we notice an relative improvement of few-shot accuracy and instant accuracy by 2.3\% and 0.4\% on two datasets. We see a similar trend on \texttt{CLIF-55}. From Figure~\ref{fig:few_shot_order_leopard}, we see \method-Reg outperforms \method-Vanilla in the default and decreasing relevance order; while we observe an outlier in \method-Reg runs in the increasing relevance order. From Figure~\ref{fig:leopard_fewshot_evolve}, we see few-shot learning accuracy improves more stable as \method-Reg learns more upstream tasks.

\paragraph{Q3: Does \method-Reg improve over HNet-Reg?} The major differences of \method-Reg compared to HNet-Reg~\cite{Oswald2020ContinualLW} are (1) few-shot task representations and (2) inferring task representations with context predictors instead of learning them as trainable embeddings. As an ablation study, we progressively replace out two components in \method~, as shown in Table~\ref{tab:hnet_abl}. We
see removing few-shot task-representation causes
the few-shot accuracy to drop on both datasets by
1.08 and 0.33 points. Using trainable task embeddings in place of task encoders causes the few-shot accuracy and final accuracy to drop on \texttt{CLIF-55}; however, on \texttt{CLIF-26}, we see a slight performance improvement. It implies task encoders are useful when the number of sequential training tasks is large (such as in \texttt{CLIF-55}). Trainable task embeddings can be an alternative in the other case\footnote{The results of trainable embedding variant have improved compared to the previous version as we fixed a bug in implementation}.



\begin{table}[tb]
\centering
\scalebox{0.65}{
\begin{tabular}{@{}lcccc@{}}
\toprule
\textbf{}       & \multicolumn{2}{c}{\textbf{\texttt{CLIF 26}}}     & \multicolumn{2}{c}{\textbf{\texttt{CLIF 55}}}  \\
\textbf{}       & \textbf{Final Acc.} & \textbf{Few-shot Acc.} & \textbf{Final Acc.} & \textbf{Few-shot Acc.} \\ 

\midrule
\method-Reg        &    $77.22_{\pm 1.1}$      &   $60.09_{\pm 1.1}$             &      $56.16_{\pm 1.6}$               &          $68.46_{\pm 0.2}$       \\
-Few-shot TR  &     $78.78_{\pm 1.3}$      & $59.01_{\pm 0.6}$      &      $55.90_{\pm 1.4}$          &   $68.13_{\pm 0.5}$         \\
+Train Embs & $77.32_{\pm 1.5}$   & $61.62_{\pm 0.1}$      &        $49.36_{\pm 1.1}$           &  $66.54_{\pm 0.2}$                  \\ \bottomrule
\end{tabular}}
\caption{Ablation study on ~\method-Reg: after removing few-shot task-representations (-Short-term TR), and replacing context predictors with trainable embeddings (+Train Embs.).}
\label{tab:hnet_abl}
\end{table}

\begin{figure}
    \centering
    \subfloat[][\texttt{CLIF-26}]{ \includegraphics[width=0.88\linewidth]{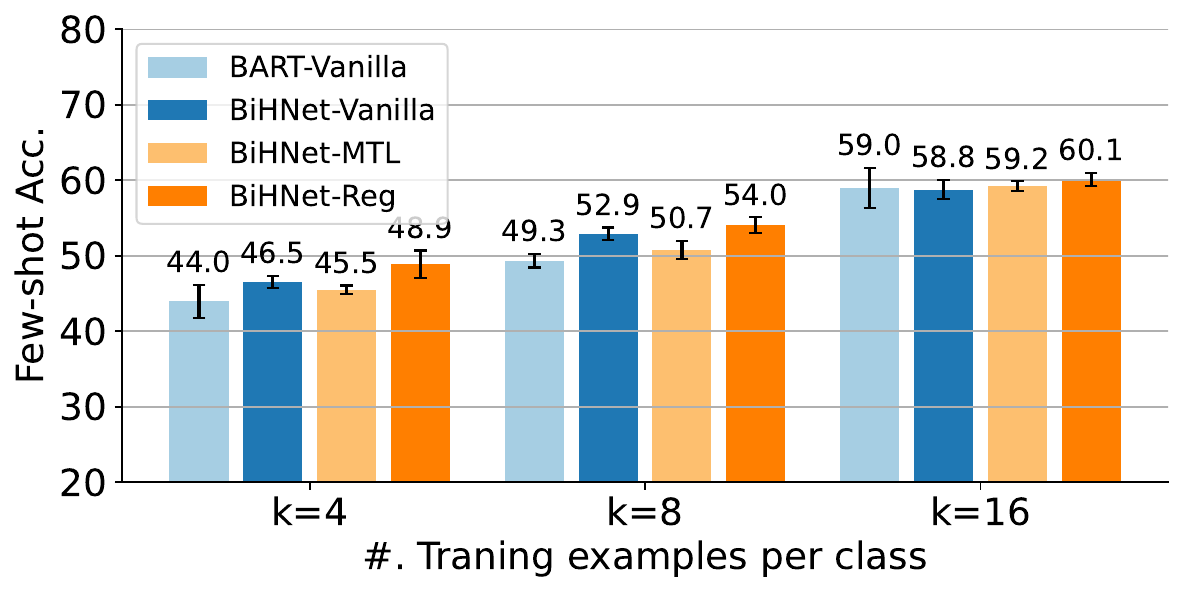}}
    
    \subfloat[][\texttt{CLIF-55}]{    \includegraphics[width=0.8\linewidth]{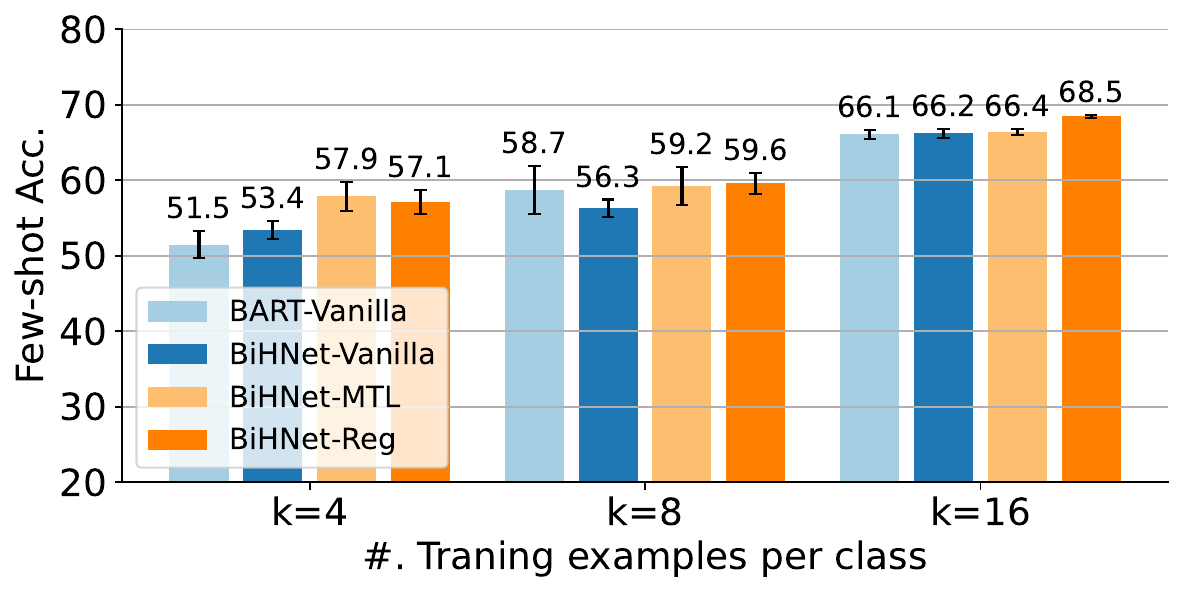}}
    \caption{Few-shot learning performance of BART-Vanilla, \method-Vanilla, \method-MTL, and \method-Reg under different number of training examples per class ($k=4,8,16$) on \texttt{CLIF-26} and \texttt{CLIF-55}. 
    }
    \label{fig:few_shot_k_leopard}
\end{figure}

\paragraph{Q4: Sensitivity Analysis: how do models perform under various number of few-shot training examples.}
Figure~\ref{fig:few_shot_k_leopard} summarizes few-shot performance of different methods under different number of training examples per class on \texttt{CLIF-26} and \texttt{CLIF-55}. We observe~\method-Reg always achieves the best performance and the improvement is generally more significant when the training sets are smaller.


\paragraph{Discussion.} Our results indicate \method-Reg could effectively improve knowledge accumulation over time compared to similar adapter learning frameworks (\method-Single and BART-Adapter-Single). However, \method-Reg does not rival BART-Single in terms or few-shot learning accuracy. We believe this is due to the restricted model capacity of adapter, as compared to fine-tuning entire transformer. 
This opens up future work on improving continual learning algorithms that are compatible with PTLM fine-tuning. 

\section{Related Work}


\paragraph{Continual Learning}
The primary challenge that is addressed in CL literature is overcoming catastrophic forgetting. Generally, existing CL methods encompass memory and generative replay-based approaches~\cite{robins1995catastrophic, lopez2017gradient,shin2017continual}, regularization based approaches~\cite{kirkpatrick2017overcoming, Nguyen2018VariationalCL} and model expansion based approaches~\cite{shin2017continual}. Recently, continual learning has drawn attention in the NLP field~\cite{Sun2020LAMOLLM, wang2019sentence, huang2021continual}.

\paragraph{Continual Meta-Learning}
There exists literature that studies continual meta-learning outside NLP application, with various definition of the problem. Some prior works~\cite{xu2019bert, dAutume2019EpisodicMI, Wang2020EfficientML} aim to develop algorithms that allows fast recovery of previous performance when a few training examples of an early task are available again at the test time.~\citet{caccia2020online} proposed a setup where models visit a sequence of potentially re-occuring tasks and measured online cumulative performance as metrics.~\citet{antoniou2020defining} assumes the model visits a sequence of few-shot classification tasks while the test tasks consist of seen classes at training. The problem setup of~\citet{Jerfel2019ReconcilingMA} is most related to ours which learns to perform few-shot learning on new tasks better, but is only studied for image classification tasks with much smaller number tasks. To our best knowledge, our work is the first to study continual knowledge accumulation for few-shot learning in diverse NLP tasks for large-scale transformer models.
\section{Conclusion}
We present the Continual Learning of Few-Shot Learners (CLIF) challenge to simulate the scenario where a learner continually accumulate (generalizable) knowledge over a sequence of NLP tasks, while retaining its performance on the seen tasks. We propose evaluation protocols to study the performance of existing continual learning algorithm, and present our method \method-Reg. We demonstrate the potentials of building a NLP system that, through continual training, can perform more tasks and also become more efficient in mastering new tasks.
Future works include extending our work to task agnostic scenarios where the distribution of data may shift continuously and studying algorithms for continual refinement of large-scale pre-trained models with emerging unlabeled data.

\section*{Acknowledgements}
This research is supported in part by the Office of the Director of National Intelligence (ODNI), Intelligence Advanced Research Projects Activity (IARPA), via Contract No. 2019-19051600007, the DARPA MCS program under Contract No. N660011924033, the Defense Advanced Research Projects Agency with award W911NF-19-20271, NSF IIS 2048211, NSF SMA 1829268, and gift awards from Google, Amazon, JP Morgan and Sony. 
We would like to thank all the collaborators in USC INK research lab for their constructive feedback on the work.

\bibliography{all_norm}
\bibliographystyle{acl_natbib}

\appendix

\clearpage

\begin{table}[]
\centering
\scalebox{0.7}{\begin{tabular}{@{}lcc@{}}
\toprule
             & Trainable Params & Total Params \\ \midrule
BART         & 139M          & 139M      \\
BART-Adapter & 72M           & 212M      \\
BiHNET         & 266M          & 405M      \\
BiHNET$_{d=4}$ & 40M           & 180M      \\ \bottomrule
\end{tabular}
}
\caption{Statistics of trainable and total model parameters in each model to learn 9 GLUE tasks.}
\label{tab:param_count}
\end{table}

\begin{table}[]
    \centering
    \scalebox{0.7}{
    \begin{tabular}{lccc}
    \toprule
    Methods & Final Acc. & Inst. Acc. & F-S Acc.    \\ \midrule
    BART-Adapter-Single & - & 74.98$_{\pm 0.7}$ & 59.00 $_{\pm 1.9}$    \\
    \method$_{d=4}$-Reg & 72.50 $_{\pm 2.3}$ & 79.45 $_{\pm 0.5}$ & 60.68 $_{\pm 1.5}$  \\ \bottomrule
    \end{tabular}
    }
    \caption{Performance when we use a smaller hidden dimension ($d$=4) for the HyperNet in \method-Reg.}
    \label{tab:small_dim}
\end{table}

\begin{table*}[th]
\centering
\scalebox{0.7}{
\begin{tabular}{ lp{19cm}c }
\toprule
 Task Order & \multicolumn{1}{c}{Tasks} \\ \midrule
\multicolumn{2}{c}{\textbf{\texttt{CLIF-26}} } \\
Default  &  cola, sst2, mrpc, qqp, stsb, mnli, qnli, wnli, rte    \\
Relevance $\downarrow$ & mnli, sst2, qqp, qnli, stsb, mrpc, cola, rte, wnli   \\
Relevance $\uparrow$ & wnli, rte, cola, mrpc, stsb, qnli, qqp, sst2, mnli     \\
\midrule
\multicolumn{2}{c}{\textbf{\texttt{CLIF-55}} }   \\
Default   &    ai2\_arc, aqua\_rat, boolq, codah, commonsense\_qa, cosmos\_qa, dream, eli5-askh, eli5-asks, eli5-eli5, freebase\_qa, hellaswag, jeopardy, kilt\_hotpotqa, kilt\_nq, kilt\_trex, kilt\_zsre, lama-conceptnet, lama-google\_re, lama-squad, lama-trex, math\_qa, mc\_taco, numer\_sense, openbookqa, qasc, quail, quarel, quartz-no\_knowledge, quartz-with\_knowledge, race-high, race-middle, sciq, search\_qa, social\_i\_qa, squad-no\_context, superglue-copa, superglue-multirc, swag, web\_questions, wino\_grande, wiqa   \\
\bottomrule
\end{tabular}
}
\caption{Order of continual learning tasks in \texttt{CLIF-26} and \texttt{CLIF-55} datasets.}
\label{tab:task_order}
\end{table*}

\section{Implementation Details}
\label{sec:apdx_imp}

We tune hyperparameters except the time steps of few-shot training on the validation set of upstream continual learning tasks. We tune the hyperpameters on \texttt{CLIF-26} and apply the same for \texttt{CLIF-55} for the same approaches. We tune learning rates by enumerating over [3e-4, 1e-4, 3e-5, 1e-5], and finally use a learning rate of 3e-5 for all MTL approaches and fine-tuend BART approaches (\eg, BART-EWC, BART-Vanilla), and a learning rate of 1e-4 for \method, HNet, and BART-Adapter-Single. We use a batch size of 64 across experiments. We train the model for at most 100 epochs for each training task with a patience of 3 epochs without validation performance improvement. Before training on a new task, we revert the model to the checkpoint with the best validation performance in the previous task. In the few-shot learning stage, we use the same learning rate and train the model for 400 epochs, assuming no validation sets to perform early stopping. The number of training steps are decided based on the performance of \method-Vanilla on airline, conll, and disaster tasks. We set the hidden size of adapters inserted between layers of BART transformers as 256 and the one in the classification head as 64. The weight generator in~\method~is implemented as a two-layer MLP with a hidden size of 32. For replay based approaches (MbPA++ and meta-MbPA), we store \textit{all} examples following these works and randomly draw mini-batches to replay every 100 training steps. For \method, HNet, and EWC, we set the regularization strength (coefficient before the regularization loss term) as 0.01 without further tuning. We use a sample size 64 to compute the few-shot task representation on \texttt{CLIF-26} and 10 for \texttt{CLIF-55} at training. Experiments are run on Nvidia Quadro 6000 or Quadro 8000 GPUs with cuda version 10.1 installed. Through out the experiments (including the hyperparameter search), we run each method with three random seeds.

\paragraph{Details of Datasets}.
For \texttt{CLIF-26}, we use the train, validation, and test split from~\citet{Bansal2020LearningTF}. For a seen trained model, we evaluate its few-shot ability over 5 different partitions of train-test splits of a single few-shot task. For \texttt{CLIF-55}, we use the train, validation, and test splits provided in the datasets library\footnote{https://huggingface.co/datasets}. The few-shot training and validation sets are random samples of the official train and validation splits; while we do not sub-sample the test split. Similarly, we evaluate few-shot learning ability over 5 different samples of training and validation examples.


\paragraph{Details of Task Orders.} Table~\ref{tab:clif55tasks} summarize the list of 45 upstream training tasks and 10 few-shot training tasks. Table~\ref{tab:task_order} further shows the order of continual learning tasks.

\section{Parameter Efficiency} 
We show the statistics of trainable and total parameters in each compared architecture in Table~\ref{tab:param_count} on \texttt{CLIF-26}. In our default settings, ~\method~has twice as many trainable parameters as BART and above three times as BART-Adapter. However, we could significantly reduce the number of parameters by setting the hidden size $d$ of the Hypernetwork smaller than the number of the tasks. We reduce $d$ to 4, and summarize the results in~\ref{tab:small_dim}. We notice the approach achieves instant accuracy and few-shot accuracy on par with \method-Reg in the standard setup. We notice the approach achieves lower final accuracy compared to the default setup, but the score is still more competitive than baselines, such as BART-MbPA and BART-meta-MbPA, and \method-Vanilla.


\begin{table*}
\centering
\scalebox{0.72}{
\begin{tabular}{l|cc}
\toprule
\textbf{Task Name} & \textbf{Task} & \textbf{Reference} \\
\midrule
\textit{Upstream tasks}  & & \\
ade\_corpus\_v2-classification &	other &	\citealt{GURULINGAPPA2012885}	\\

circa &	other &	\citealt{louis-etal-2020-id}	\\

discovery &	other &	\citealt{sileo-etal-2019-mining}	\\

emotion &	emotion & \citealt{saravia-etal-2018-carer}	\\
ethos-directed\_vs\_generalized &	hate speech detection &	\citealt{Mollas2020ETHOSAO}	\\
ethos-disability &	hate speech detection &	\citealt{Mollas2020ETHOSAO}	\\
ethos-gender &	hate speech detection &	\citealt{Mollas2020ETHOSAO}	\\
ethos-sexual\_orientation &	hate speech detection &	\citealt{Mollas2020ETHOSAO}	\\

glue-cola &	other &	\citealt{warstadt-etal-2019-neural}	\\
glue-mnli &	nli & \citealt{williams-etal-2018-broad}	\\
glue-mrpc &	paraphrase & \citealt{dolan-brockett-2005-automatically}\\
glue-qnli &	nli &	\citealt{rajpurkar-etal-2016-squad}	\\
glue-qqp &	paraphrase & \href{http://data.quora.com/First-Quora-Dataset-Release-Question-Pairs}{(link)}	\\
glue-rte &	nli &	\begin{tabular}[c]{@{}l@{}}\citealt{dagan2005pascal, bar2006second}\\\citealt{giampiccolo2007third, bentivogli2009fifth}\end{tabular}	\\
glue-sst2 &	sentiment analysis &	\citealt{socher-etal-2013-recursive}	\\
glue-wnli &	nli & \citealt{levesque2012winograd}	\\
google\_wellformed\_query &	other &	\citealt{faruqui-das-2018-identifying}	\\
hate\_speech\_offensive &	hate speech detection &	\citealt{hateoffensive}	\\
hatexplain &	hate speech detection &	\citealt{mathew2020hatexplain}	\\
health\_fact &	fact checking &	\citealt{kotonya-toni-2020-explainable-automated}	\\
imdb &	sentiment analysis & \citealt{maas-etal-2011-learning}	\\
kilt\_fever &	fact checking &	\citealt{thorne-etal-2018-fever}	\\
liar &	fact checking &	\citealt{wang-2017-liar}	\\
onestop\_english &	other &	\citealt{vajjala-lucic-2018-onestopenglish}	\\
paws &	paraphrase & \citealt{zhang-etal-2019-paws}	\\
rotten\_tomatoes &	sentiment analysis & \citealt{pang-lee-2005-seeing}	\\
scicite &	other &	\citealt{cohan-etal-2019-structural}	\\
scitail &	nli & \citealt{scitail}	\\
sick &	nli &	\citealt{marelli-etal-2014-sick}	\\
sms\_spam &	other &	\citealt{sms_spam}	\\

superglue-rte &	nli & \begin{tabular}[c]{@{}l@{}}\citealt{dagan2005pascal, bar2006second}\\\citealt{giampiccolo2007third, bentivogli2009fifth}\end{tabular}	\\
superglue-wic &	other &	\citealt{pilehvar-camacho-collados-2019-wic}	\\
superglue-wsc &	other &	\citealt{levesque2012winograd}	\\

trec &	other &	\citealt{li-roth-2002-learning,hovy-etal-2001-toward}	\\
trec-finegrained &	other &	\citealt{li-roth-2002-learning,hovy-etal-2001-toward}	\\
tweet\_eval-emoji &	emotion & \citealt{barbieri-etal-2020-tweeteval}	\\
tweet\_eval-emotion &	emotion &	\citealt{barbieri-etal-2020-tweeteval}	\\
tweet\_eval-irony &	emotion &	\citealt{barbieri-etal-2020-tweeteval}	\\
tweet\_eval-offensive &	emotion &	\citealt{barbieri-etal-2020-tweeteval}	\\
tweet\_eval-sentiment &	emotion &	\citealt{barbieri-etal-2020-tweeteval}	\\
tweet\_eval-stance\_abortion &	emotion &	\citealt{barbieri-etal-2020-tweeteval}	\\
tweet\_eval-stance\_climate &	emotion &	\citealt{barbieri-etal-2020-tweeteval}	\\
tweet\_eval-stance\_hillary &	emotion &	\citealt{barbieri-etal-2020-tweeteval}	\\
wiki\_auto &	other &	\citealt{jiang-etal-2020-neural}	\\

yahoo\_answers\_topics &	topic &	\href{https://webscope.sandbox.yahoo.com/catalog.php?datatype=l}{(link)}	\\

\textit{Few-shot learning tasks}  & & \\
superglue-cb &	nli & \citealt{Marneffe_Simons_Tonhauser_2019}	\\
dbpedia\_14 &	topic &	\citealt{Lehmann2015DBpediaA}	\\
wiki\_qa &	other &	\citealt{yang-etal-2015-wikiqa}	\\
emo &	emotion & \citealt{chatterjee-etal-2019-semeval}	\\
yelp\_polarity &	sentiment analysis & \citealt{zhang2015character}; \href{https://www.yelp.com/dataset}{(link)}	\\
ethos-religion &	hate speech detection &	\citealt{Mollas2020ETHOSAO}	\\
financial\_phrasebank &	sentiment analysis &	\citealt{financial-phrasebank}	\\
tab\_fact &	fact checking &	\citealt{Chen2020TabFact}	\\
anli &	nli & \citealt{nie-etal-2020-adversarial}	\\
ethos-race &	hate speech detection &	\citealt{Mollas2020ETHOSAO}	\\

\bottomrule
\end{tabular}
}
\caption{Datasets and tasks included in \texttt{CLIF-55} for upstream training and few-shot learning.}
\label{tab:clif55tasks}
\end{table*}



\end{document}